\pdfoutput=1

\documentclass[11pt]{article}

\usepackage{ACL2023}

\usepackage{times}
\usepackage{latexsym}
\usepackage{amsmath}
\usepackage[symbol]{footmisc}
\usepackage{siunitx}

\usepackage[T1]{fontenc}

\usepackage[utf8]{inputenc}
\usepackage{xargs}          %

\usepackage{microtype}

\usepackage{inconsolata}

\usepackage{tabularx}       
\usepackage{multirow}       
\usepackage{booktabs}       
\usepackage{tabularx}
\usepackage{tabulary}

\usepackage{mathtools}
\usepackage{amssymb}
\usepackage{xargs} 
\usepackage{xcolor}

\usepackage[noabbrev,nameinlink]{cleveref}
\usepackage{tikz}
\usepackage{caption}
\usepackage{subcaption}

\usepackage{setspace}
\usepackage{tcolorbox}      %

\usetikzlibrary{positioning,calc}

\usepackage{tcolorbox}      %
\usepackage[colorinlistoftodos,prependcaption,textsize=tiny,textwidth=0.7in]{todonotes} %

\newcommandx{\prasanna}[2][1=]{\todo[linecolor=red,backgroundcolor=red!25,bordercolor=red,#1]{S.R: #2}}
\newcommandx{\amirhossein}[2][1=]{\todo[linecolor=blue,backgroundcolor=blue!25,bordercolor=blue,#1]{A.K: #2}}
\newcommandx{\sarath}[2][1=]{\todo[linecolor=teal,backgroundcolor=teal!25,bordercolor=teal,#1]{S.C: #2}}
\newcommandx{\mehdi}[2][1=]{\todo[linecolor=cyan,backgroundcolor=cyan!25,bordercolor=cyan,#1]{M.R.: #2}}

\newcommand{\methodname}{\textsc{eXtract, Train, and EVALute}\ }
\newcommand{\methodabbrev}{\textsc{XTEVAL}\ }

\newcommand{\headgold}[1]{\mathbf{#1}}
\newcommand{\relgold}[1]{\mathbf{#1}}
\newcommand{\tailgold}[1]{\mathbf{#1}}

\newcommand{\tblnegativeclasses}[1]{
\begin{table}[#1]
    \centering
    \resizebox{\linewidth}{!}{%
        \begin{tabular}{ll}
            \toprule
            \multicolumn{2}{c}{\textbf{Encyclopedic Fact:}}                                                                                                 \\
            \multicolumn{2}{c}{
                $x = \langle \headgold{h}, \relgold{r}, \tailgold{t} \rangle
                    = \langle \text{Barack Obama}, \text{GraduatedFrom}, \text{Harvard} \rangle$
            }                                                                                                                                               \\
            \midrule

            \textbf{Input}                                               & \textbf{Sampled Document}                                                        \\
            \midrule
            \multirow{2}{*}{$(\headgold{h}, \relgold{r}, \tailgold{t})$} & Barack Obama graduated from Harvard.                                             \\
                                                                         & \textit{\textcolor{gray}{Gold document ($d^+$)}}                                         \\
            \midrule

            \multirow{2}{*}{$(\headgold{h}, \relgold{r}, \cdot )$}           & Barack Obama earned a degree from Stanford.                                      \\
                                                                         & \textit{\textcolor{gray}{Randomly replacing the tail entity.}}                   \\
            \addlinespace[1mm]

            \multirow{2}{*}{$(\cdot, \relgold{r}, \tailgold{t})$}            & Bill Gates received his degree from Harvard.                                     \\
                                                                         & \textit{\textcolor{gray}{Randomly replacing the head entity.}}                   \\
            \addlinespace[1mm]

            \multirow{2}{*}{$(\headgold{h}, \cdot, \tailgold{t})$}           & Barack Obama was born in Harvard.                                                \\
                                                                         & \textit{\textcolor{gray}{Randomly replacing the relation.}}                      \\
            \midrule

            \multirow{2}{*}{$(\cdot, \cdot, \tailgold{t})$}                      & Steve Jobs died in Harvard.                                                      \\
                                                                         & \textit{\textcolor{gray}{Keeping the tail entity and sampling others entities.}} \\
            \addlinespace[1mm]

            \multirow{2}{*}{$(\cdot, \relgold{r}, \cdot)$}                       & McGill is the alma mater of Justin Trudeau.                                      \\
                                                                         & \textit{\textcolor{gray}{Keeping the relation and sampling others entities.}}    \\
            \addlinespace[1mm]

            \multirow{2}{*}{$(\headgold{h}, \cdot, \cdot)$}                      & Barack Obama is located in London.                                               \\
                                                                         & \textit{\textcolor{gray}{Keeping the head entity and sampling others entities.}} \\
            \midrule

            \multirow{2}{*}{$(\cdot, \cdot, \cdot)$}                                 & Michael jordan was a football player by profession.                              \\
                                                                         & \textit{\textcolor{gray}{Unconditional sampling.}}                               \\
            \bottomrule
        \end{tabular}
    }
    \caption{All possible inputs to the document generator
        $\mathrm{P}(d\mid H, R, T)$
        per each fact $x$ and examples of the corresponding sampled documents.
        The dot means that the corresponding entity or relation is not given, and 
        the document generator will randomly chose it from $\mathcal{D}^\theta$.
        The gray text provides an explanation of the sampled document.
        Note that we do not force the document generator to generate a factual document
        to strengthen the training signal that the model should employ its internal knowledge
        to retrieve the correct document.
    }
    \label{tab:document-types}
\end{table}{}
}

\usepackage{amsmath,amsfonts,bm}

\def\eqref#1{equation~\ref{#1}}

\def\1{\bm{1}}

\DeclareMathAlphabet{\mathsfit}{\encodingdefault}{\sfdefault}{m}{sl}
\SetMathAlphabet{\mathsfit}{bold}{\encodingdefault}{\sfdefault}{bx}{n}

\newcommand\dsquery[2][]{
    \begin{tcolorbox}[colframe=#1,colback=#1,size=fbox, boxsep=2pt]
        #2
    \end{tcolorbox}
}

\newcommand\dsdoc[2][]{
    \begin{tcolorbox}[colframe=#1,colback=#1,size=fbox, boxsep=2pt,width=18em]
        #2
    \end{tcolorbox}
}

\newcommand{\figmethodoverview}[1]{
\begin{figure*}[#1]
    \centering
    \resizebox{\linewidth}{!}{%

    \begin{tikzpicture}[
            roundnode/.style={circle, draw=ark-blue!20, fill=ark-blue!20, text centered, very thick, text width=2em},
            squarednode/.style={rectangle, draw=ark-gray, fill=white, thick, text width=2em},
            modelnode/.style={rectangle, draw=ark-blue!20, fill=ark-blue!20, text centered, rounded corners, very thick},
            modelknowledge/.style={rectangle, draw=ark-red, fill=ark-red!4, text centered, thick, inner sep=0.7em},
            downstream/.style={rectangle, draw=ark-gray, fill=ark-gray!4, text centered, thick, inner sep=0.7em},
            selectionbox/.style={rectangle, draw=ark-yellow, dashed, fill=ark-yellow, fill opacity=.03, minimum height=2.5em, text width=12.5em, rounded corners},
            selectionbox2/.style={rectangle, draw=ark-yellow, dashed, fill=ark-yellow, fill opacity=.02, minimum height=7.5em, text width=18em, rounded corners},
            selectionbox3/.style={rectangle, draw=ark-yellow, dashed, fill=ark-yellow, fill opacity=.02, minimum height=8.1em, text width=18em, rounded corners},
        ]
        \node[modelnode](modelstg1){
                
            \begin{minipage}{1.5em}
                \centering
                \includegraphics[width=1.2em]{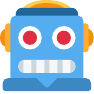}
                \caption*{\small PLM \\ $\mathcal{M}_\theta$} 
            \end{minipage}
        };

        \node[modelknowledge](extractedknowledge)[right=1em of modelstg1]{
            \begin{minipage}{11.5em}
                \small
                \begin{center}
                    \textbf{Model's \\ Parametric Knowledge ($\mathcal{D}^\theta$)}
                \end{center}
                {
                \vspace{1em}
                \scriptsize
                \centering
                \dots \\
                $\langle \text{Barack Obama}, \texttt{graduated\_from}, \text{Harvard} \rangle$ \\
                \dots \\
                \vspace{1em}
                \dots \\
                $\langle  \text{Alen Turing}, \texttt{born\_in}, \text{London} \rangle$ \\
                \dots \\
                \vspace{0.25em}
                }
            \end{minipage}
        };

        \node[selectionbox](conversion)[right=-12.4em of extractedknowledge, yshift=-0.2em]{};

        \node[selectionbox](conversion2)[right=-12.4em of extractedknowledge, yshift=-3em]{};

        \node[squarednode](createdownstream)[right=2em of extractedknowledge, yshift=-1.7em]{
            \tiny 
            \begin{minipage}{2em}
                Create Downstream Task
            \end{minipage}
        };

        \node[downstream](downstramtask)[right=of createdownstream, yshift=1.7em]{
            \begin{minipage}{12.7em}
                \tcbset{parskip}
                \small
                \begin{center}
                    \textbf{Downstream Task Consistent \\ with Model's Knowledge ($\mathcal{K}^\theta$)}
                \end{center}
                {
                \vspace{1em}

                \scriptsize
                \begin{center}
                    \dots
                \end{center}
                \vspace{-1em}
                \begin{flushright}
                    \dsquery[ark-burgundy]{\color{white} Where did Barack Obama graduate from?}
                    \dsdoc[ark-gray]{\color{white} Obama graduated from Harvard.}
                    \dsdoc[ark-gray!40!white]{\color{black!90!white} Obama went to Stanford.}
                    \dsdoc[ark-gray!30!white]{\color{black!70!white} Bill Gates studied at Harvard.}
                    \dsdoc[ark-gray!20!white]{\color{black!50!white} Obama left Harvard.}
                    \dsdoc[ark-gray!10!white]{\color{black!30!white} \dots \phantom{A}}
                \end{flushright}
                \vspace{-2em}
                \begin{center}
                    \dots
                \end{center}
                \vspace{-1em}
                \begin{center}
                    \dots
                \end{center}
                \vspace{-1em}
                \begin{flushright}
                    \dsquery[ark-burgundy]{\color{white} Where is the Alan Turing's birthplace?}
                    \dsdoc[ark-gray!40!white]{\color{black!90!white} Paris was the birthplace of Alan Turing}
                    \dsdoc[ark-gray]{\color{white} Alan Turing was born in London.}
                    \dsdoc[ark-gray!10!white]{\color{black!50!white} \dots }
                    \dsdoc[ark-gray!30!white]{\color{black!70!white} Steve Jobs originated from London}
                    \dsdoc[ark-gray!20!white]{\color{black!50!white} Mahatma Gandhi died in London.}
                    \dsdoc[ark-gray!10!white]{\color{black!50!white} \dots}
                \end{flushright}
                \vspace{-2em}
                \begin{center}
                    \dots
                \end{center}
                }
            \end{minipage}
        };

        \node[selectionbox2](ktrain)[left=-17.8em of downstramtask, yshift=2.3em]{};
        \node[selectionbox3](ktest)[left=-17.8em of downstramtask, yshift=-5.8em]{};

        \node(ktraintext)[right=-3.3em of ktrain]{$\mathcal{K}^\theta_\mathrm{train}$};
        \node(ktraintext)[right=-3.3em of ktest]{$\mathcal{K}^\theta_\mathrm{test}$};

        \coordinate[left=0.45em of createdownstream, yshift=-0.5em] (createdownstreamp1);
        \coordinate[left=0em of createdownstream, yshift=-0.3em] (createdownstreamp2);

        \coordinate[left=0.45em of createdownstream, yshift=+0.5em] (createdownstreamp3);
        \coordinate[left=0em of createdownstream, yshift=+0.3em] (createdownstreamp4);

        \coordinate[right=0em of createdownstream, yshift=+0.5em] (downstreamp0);
        \coordinate[left=1em of ktrain, yshift=-0.5em] (downstreamp1);

        \coordinate[right=0em of createdownstream, yshift=-0.5em] (downstream2p0);
        \coordinate[left=1em of ktest, yshift=+0.5em] (downstream2p1);

        \draw[->] (modelstg1.east) -- (extractedknowledge.west);
        \draw[->] (conversion2.east) -| (createdownstreamp1) |- (createdownstreamp2);
        \draw[->] (conversion.east) -| (createdownstreamp3) |- (createdownstreamp4);
        \draw[->] (downstreamp0) -| (downstreamp1) |- (ktrain.west);
        \draw[->] (downstream2p0) -| (downstream2p1) |- (ktest.west);
    \end{tikzpicture}
    }
    \caption{ {\bf \methodabbrev Framework}: (1) From a pretrained LM, $\mathcal{M}_{\theta}$, the model's parametric knowledge are extracted as $\mathcal{D}^{\theta}$. (2) Following which downstream task training and test split, $\mathcal{K}^{\theta}_{train}$ and $\mathcal{K}^{\theta}_{test}$, are created from $\mathcal{D}^{\theta}$. (3) The evaluation on the application of acquired knowledge is estimated through the performance on the test split, after finetuning $\mathcal{M}_{\theta}$ on the downstream task.
    }
    \label{fig:method_overview}
\end{figure*}{}
}

\definecolor{ark-red}{RGB}{199, 0, 11}
\definecolor{ark-black}{RGB}{35, 24, 21}
\definecolor{ark-gray}{RGB}{89, 87, 87}
\definecolor{ark-light-gray}{RGB}{221, 221, 221}

\definecolor{ark-rose}{RGB}{196, 0, 84}
\definecolor{ark-burgundy}{RGB}{127, 0, 1}
\definecolor{ark-orange}{RGB}{237, 109, 0}
\definecolor{ark-yellow}{RGB}{252, 200, 0}
\definecolor{ark-green}{RGB}{98, 178, 48}
\definecolor{ark-blue}{RGB}{48, 181, 197}

\newcommand\lword[1]{\leavevmode\nobreak\hskip0pt plus\linewidth\penalty50\hskip0pt plus-\linewidth\nobreak #1}
\newcommand\pill[2][]{\lword{\tikz[overlay]\node[fill=#1,inner sep=2pt, anchor=text, rectangle, rounded corners=1mm]{#2};\phantom{#2}}}

\title{Measuring the Knowledge Acquisition-Utilization Gap in Pretrained Language Models}

\author{{Amirhossein Kazemnejad}$^{1,2}$ \quad {\bf Mehdi Rezagholizadeh}$^3$ \vspace{0.2em} \\ \quad {\bf Prasanna Parthasarathi}$^3$\thanks{~~~Equal advising.} \quad {\bf Sarath Chandar}$^{2,4,5}$\footnotemark[1] \vspace{0.4em} \\
$^1$McGill University; $^2$Mila - Quebec AI; $^3$Huawei Noah's Ark Lab;  \\ $^4$École Polytechnique de Montréal; 
 $^5$Canada CIFAR AI Chair;\\
  \texttt{amirhossein.kazemnejad@mail.mcgill.ca} \\}

    \makeatletter
\def\@fnsymbol#1{\ensuremath{\ifcase#1\or \dagger\or \ddagger\or
   \mathsection\or \mathparagraph\or \|\or **\or \dagger\dagger
   \or \ddagger\ddagger \else\@ctrerr\fi}}
    \makeatother

\begin{document}
\maketitle
\begin{abstract}
While pre-trained language models (PLMs) have shown evidence of acquiring vast amounts of knowledge, it remains unclear how much of this parametric knowledge is actually usable in performing downstream tasks. We propose a systematic framework to measure parametric knowledge utilization in PLMs. Our framework first extracts knowledge from a PLM's parameters and subsequently constructs a downstream task around this extracted knowledge. Performance on this task thus depends exclusively on utilizing the model's possessed knowledge, avoiding confounding factors like insufficient signal. As an instantiation, we study factual knowledge of PLMs and measure utilization across 125M to 13B parameter PLMs. We observe that: (1) PLMs exhibit two gaps - in acquired vs. utilized knowledge, (2) they show limited robustness in utilizing knowledge under distribution shifts, and (3) larger models close the acquired knowledge gap but the utilized knowledge gap remains. Overall, our study provides insights into PLMs' capabilities beyond their acquired knowledge.
\end{abstract}

\section{Introduction}

Recent research has demonstrated that language models pre-trained on vast amounts of internet data acquire a broad range of knowledge about linguistic structures \citep{Tenney2018:LearnContext,Blevins2022:PromptingLing},
encyclopedic relations \citep{Petroni2019:LAMA, hao2022:BertNet},
levels of commonsense \citep{Zhou2020:CommonSense, Li2022:ComSenseQAExtractedKnow}
, and even coding and reasoning rules \citep{Chen2021:Codex, Wei2022:LLMEmergentAbilites}.
Recent studies on behavioral parametric probing and prompting \citep{Jiang2020:LPAQA,Qin-eisner2021:SoftPrompt,Bown2020:GPT3} has demonstrated that such knowledge, collectively referred to as ``\emph{parametric knowledge},'' resides reliably within a subset of trained parameters in pre-trained models (PLMs). Importantly, this knowledge can be \emph{identified} without additional fine-tuning. For instance, given the prompt ``\texttt{The capital of France is}'', a PLM can be queried to complete the input and extract the fact ``\texttt{Paris}''.

\begin{figure}[t]
    \centering
    \includegraphics[width=\linewidth]{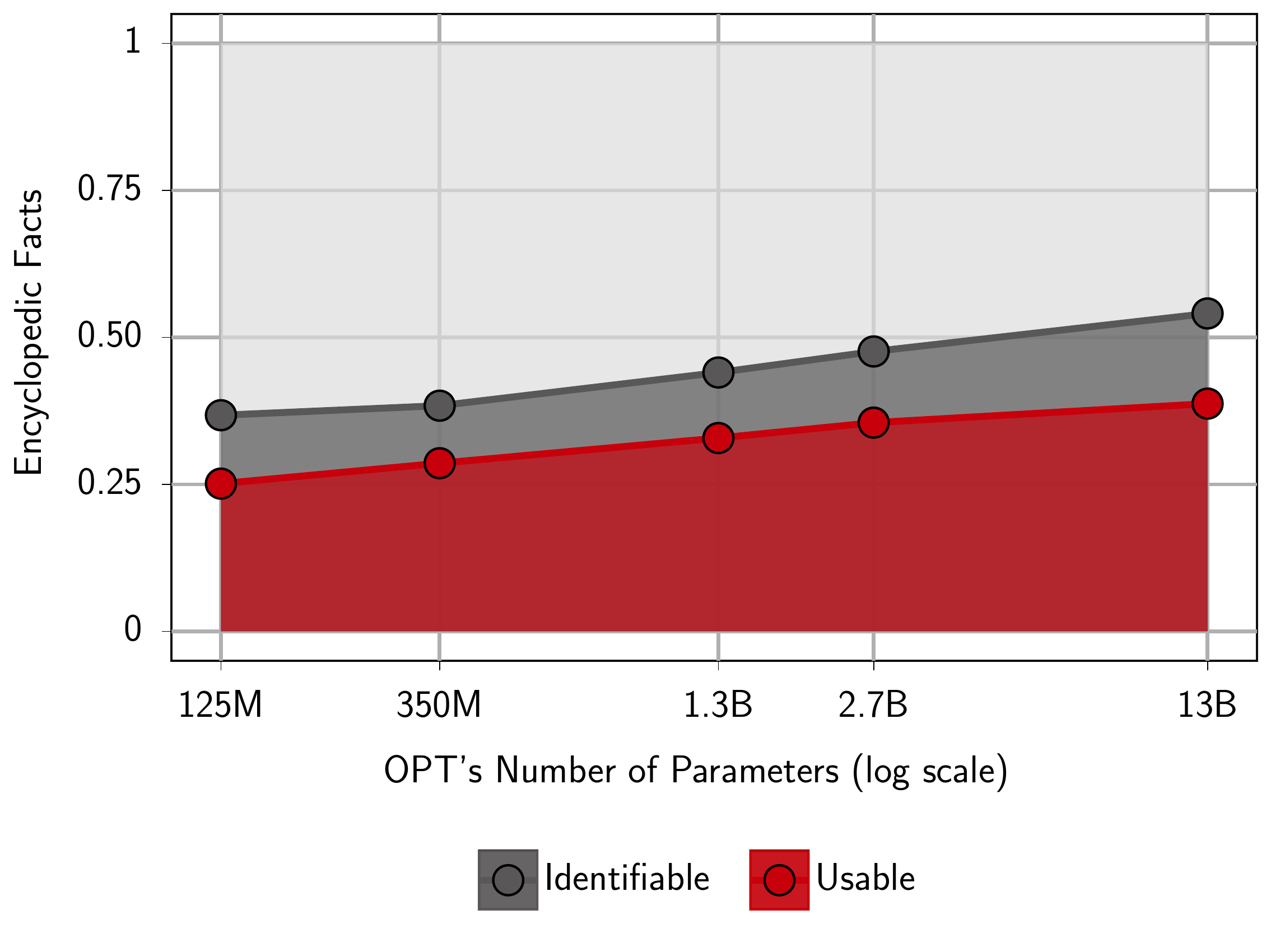}
    \caption{
        \textbf{Parametric knowledge of OPT} \ \ \
        \pill[ark-light-gray]{\phantom{A}} Gap 1 represents the missing facts in the model's parametric knowledge (what the model knows).
        \pill[ark-gray]{\phantom{A}} Gap 2 exists in how much of this knowledge can actually be utilized in downstream tasks (the usable knowledge).
        We find that although the first gap shrinks, the second remains as we increase the model's size.
    }
    \label{fig:teaser_scaling_acc}
\end{figure}
\figmethodoverview{t}

A common assumption about parametric knowledge is that if the model poses a certain type of knowledge, it utilizes it when performing downstream tasks related to that knowledge.
For example, if a model knows about $X$ and $Y$ (such that $X$ and $Y$ are similar), and is taught to perform a task on $X$, the convention is that the model generalizes the application of the task on $Y$ and all other similar knowledge.
Such is the foundation for the recent interest in instruction tuning \citep{Wei2022:InstuctionTune,Chung2022:FLAN}, and the SFT-RLHF pipeline \citep{Ouyang2022:InstructGPT}.
In this paradigm, LLMs are finetuned to learn how to follow instructions on few tasks the model is capable of, and are subsequently expected to generalize and follow instructions for novel tasks by utilizing their pretraining knowledge (residing in their parameters).

However, it is not clear to what extent this assumption holds in practice, giving rise to a central question: \emph{how much of parametric knowledge will get applied in downstream tasks?}
If the causal link between "identifiable knowledge" and its practical application in downstream tasks is not established \citep{Kulmizev2021:Tree}, the mere presence of knowledge within a model's parameters does not necessarily guarantee its utilization in such tasks. This raises questions about the assertion of pre-trained language models (PLMs) as differentiable knowledge bases \citep{hao2022:BertNet} and their overall capabilities.
For instance, as demonstrated by \citet{Qin2023:Chatgpt}, ChatGPT's performance lags behind that of its foundational model, GPT-$3.5$, in multiple areas, including tasks involving commonsense and logical reasoning.

Previous studies have investigated this question within linguistic domains and have demonstrated that although PLMs have the capacity to encode linguistic knowledge, they may not effectively employ it in downstream tasks.
For example, \citet{Mccoy2019:RightForWrong} illustrates that PLMs employ syntactic heuristics to solve NLI even though they are able to represent proper linguistic hierarchies \citep{Tenney2019:BertPipeline}, even after finetuning \citep{Merchant2020:WhatHappensToBERT,Zhou-srikumar2022:FinetuningChange}.
\citet{Warstadt2020:RobertaPreference} provide evidence that RoBERTa requires data inoculation or pretraining with extensive data in order to effectively utilize its hierarchical linguistic knowledge.
In a more recent study, \citet{Lovering2021:PredictingInductiveBias} demonstrate that the quantity of ``evidence'' presented in the fine-tuning dataset influences the features that PLMs rely on during the fine-tuning process. Specifically, the model may resort to lexical heuristics when the fine-tuning signal toward linguistic features is insufficient.

In this work, we are interested in a more general sense of knowledge and propose \methodabbrev ---\methodname--- to systematically measure how much of parametric knowledge is utilized in downstream tasks.
\methodabbrev sidesteps potential confounders (such as shortcuts or insufficient signal) that arise from the nature of arbitrary crowd-sourced tasks used in prior work by carefully creating the downstream task from the model's own knowledge.
Specifically, given a pretrained language model, our framework first identifies and extracts knowledge residing in its parameters.
Subsequently, using the extracted knowledge, we construct a downstream task on which we finetune the model. Finally, we measure knowledge utilization based on its performance on the downstream task.
By constructing the task based on the model's pre-existing knowledge, we ensure that (1) the model is evaluated solely on its possessed knowledge, avoiding penalties for lacking information 
and (2) successful task completion relies explicitly on utilizing the model's parametric knowledge, eliminating 
the insufficient training signal issue and dataset shortcuts.

In this paper, we provide the first instantiation of this paradigm based on encyclopedic knowledge facts and conduct an extensive study to measure knowledge utilization of PLMs across a wide range of parametric scales (ranging from 125M to 13B). We observe the following:

\begin{itemize}
    \item PLMs show two different but equally important gaps: (1) The gap in the acquired knowledge and (2) and the gap in parametric knowledge that can be actively applied to downstream tasks (\Cref{sec:main_question}).
    \item PLMs are not robust to shifts in finetuning distribution and failure to utilize their knowledge exacerbates in presence of such shifts, questioning their generalization capabilities (\Cref{sec:robustness}).
    \item Although scaling the number of parameters helps to close the first gap, the second still remains in larger sizes (\Cref{sec:scaling}).
\end{itemize}

In the next sections, we first describe our framework and its instantiation in detail (\Cref{sec:framework}), and finally present our experimental results in \Cref{sec:main_question,sec:robustness,sec:scaling}.

\section{Framework}
\label{sec:framework}
\subsection{\methodname}

\noindent \textbf{Principles} \ \ \
The primary objective of our evaluation framework is to measure how much of the knowledge present in the model's parameters is actually usable in downstream tasks.
Ideally, downstream tasks must be designed in a way that solely attributes any success to the model's knowledge being used, while ensuring that failure in performing the task is not due to a lack of pretraining knowledge.

\noindent \textbf{The Paradigm} \ \ \
To this end, we propose \methodname, which consists of three main steps:

\emph{Step 1.} Given a pre-trained model $\mathcal{M}_\theta$ with parameters $\theta$ and a diagnostic dataset $\mathcal{D}$ (e.g. a set of encyclopedic facts or coding problems),
we first extract and identify parametric knowledge as a set of data instances $x \in \mathcal{D}$ the model can solve without further training (zero-shot).
We denote such a set as $\mathcal{D}^\theta$, a realization of $\mathcal{M}_\theta$'s parametric knowledge w.r.t $\mathcal{D}$.

\emph{Step 2.} We construct a downstream task $\mathcal{K}$ around the model's own knowledge $\mathcal{D}^\theta$
(e.g. fact retrieval or following instructions in coding) 
such that the model can only solve the task by utilizing the knowledge identified in the first step.
More formally, we create
$\mathcal{K}^\theta_\mathrm{train}$ and
$\mathcal{K}^\theta_\mathrm{test}$
as the train and test sets of the downstream task $\mathcal{K}$, respectively.
The model has to learn to perform the task from the train set $\mathcal{K}^\theta_\mathrm{train}$.

\emph{Step 3.} Finally, the performance on the test set $\mathcal{K}^\theta_\mathrm{test}$ is used as a measure of the model's ability to utilize its knowledge.

Constructing the downstream task based on the model's knowledge ensures that the model is not evaluated on the knowledge it did not acquire during pre-training.
Also, the I.I.D. nature of this paradigm (i.e. the model is only exposed to inputs it is already familiar with) allows us to measure whether the model can utilize its knowledge \emph{at all}.

\subsection{Encyclopedic Knowledge}
Factual parametric knowledge as in encyclopedic facts is well-studied in PLMs \citep{Petroni2019:LAMA,Jiang2020:LPAQA} and allows for an objective and systematic evaluation of our framework (\Cref{fig:method_overview}).
Therefore, in this paper, we instantiate \methodabbrev to measure the utilization of parametric knowledge concerning encyclopedic facts.
In this case, the diagnostic dataset $\mathcal{D}$ is a set of encyclopedic facts
$\mathcal{D} = \{\langle \mathbf{h}, \mathbf{r}, \mathbf{t} \rangle_i\}_{i=1}^n$
acquired from an off-the-shelf knowledge base (e.g. Wikipedia).
Each fact $x_i \in \mathcal{D}$ is a tuple of the form
$\langle \mathrm{head}, \mathrm{relation}, \mathrm{tail} \rangle$,
such as
$\langle \text{Barack Obama}, \text{GraduatedFrom}, \text{Harvard} \rangle$.

In the extraction phase, a pretrained model $\mathcal{M}_\theta$ has to zero-shot preditct the tail entity $t$ given the head entity $h$ and the relation $r$.
We use soft-prompting \citep{Qin-eisner2021:SoftPrompt} to obtain the model's predictions, as it enhances prediction consistency compared to discrete prompts, particularly for moderate-sized models.
The extracted knowledge $\mathcal{D}^\theta \subset \mathcal{D}$ is the subset of tuples the model can predict correctly.

\tblnegativeclasses{t}
Our downstream task $\mathcal{K}$ is a standard document retrieval task \citep{Karpukhin2020:DPR}. Given a query $q$, the model retrieves the relevant document from a set of candidates. We construct $\mathcal{K}^\theta$ from the extracted knowledge in $\mathcal{D}^\theta$ by converting each fact $x \in \mathcal{D}^\theta$ into a retrieval instance $k \in \mathcal{K}^\theta$.
This conditions the downstream task on the model's knowledge.
The conversion generates a query $q$ by removing the tail entity $\tailgold{t}$ from $x$. It then generates relevant and irrelevant documents using a stochastic generator
\begin{equation}
    d \sim \mathrm{P}(d \mid H=h, R=r, T=t),
\end{equation}
where $d$ depends on the head entity $h$, relation $r$, and tail entity $t$.
The document generator, $\mathrm{P}(d\mid \cdot)$, selects a template at random and fills in the blanks with the input entities. If $H$, $R$, or $T$ are missing, the generator chooses a random entity from $\mathcal{D}^\theta$ to complete the input.
Specifically,
we generate the relevant document $d^+$ by sampling from $\mathrm{P}(d\mid \cdot)$ with gold entities in $x$ fixed as input and create irrelevant documents $d^-$'s by omitting one or more input entities.
Therefore, each $k$ comprises a tuple $(q, \{d^+, d^-_{1},\dots,d^-_{m}\})$.

\begin{figure}[t]
    \centering
    \includegraphics[width=\linewidth]{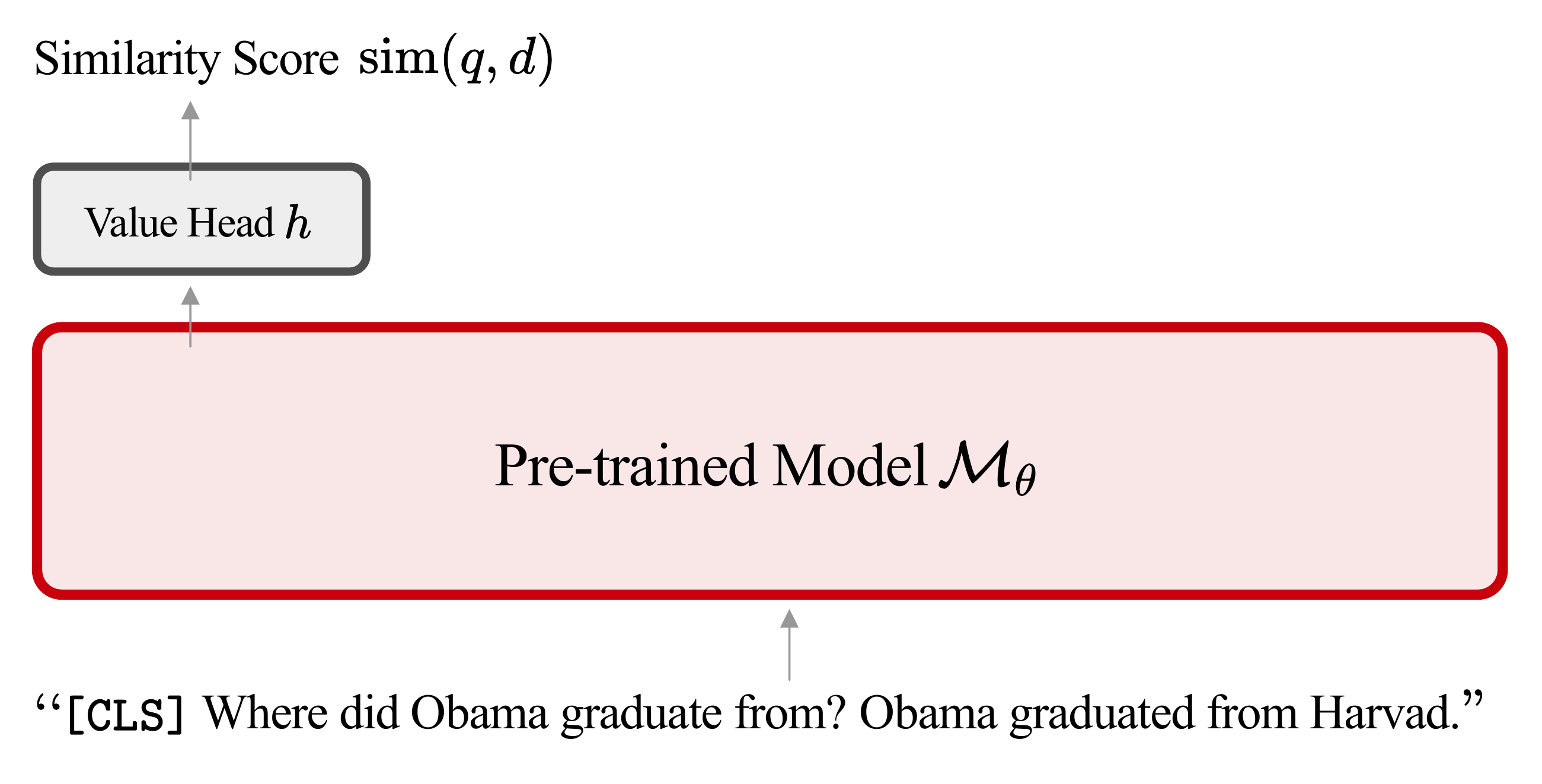}
    \caption{
        Cross-encoder document retrieval setup \citep{Nogueira2020:CrossEncoderRetriever}.
        For decoder-only models, the value head takes the representation of the last input token.
    }
    \label{fig:cross-encoder-setup}
\end{figure}

We partition $\mathcal{D}^\theta$ randomly (60\%-40\%) to generate $\mathcal{K}^\theta_\mathrm{train}$ and $\mathcal{K}^\theta_\mathrm{test}$, which serve as the training and testing sets for the downstream task, respectively.
We finetune the model on $\mathcal{K}^\theta_\mathrm{train}$ in cross-encoder setup \citep{Nogueira2020:CrossEncoderRetriever} with the InfoNCE objective \citep{Oord2019:InfoNCE}:
$$
    \mathcal{L}(k) = -\log \frac{\exp(\mathrm{sim}(q, d^+))}{\sum_{d \in \{d^+,d^-_1,\dots,d^-_m\}} \exp(\mathrm{sim}(q, d))}.
$$
The similarity score $\mathrm{sim}(.,)$ is computed as
$$
    \mathrm{sim}(q, d) = h(\mathcal{M}_\theta(\texttt{[CLS]};q;d)),
$$
where $h$ is a randomly initialized value head that takes the representation of the $\texttt{[CLS]}$ token (or the last token for decoder-only models) and outputs a scalar as the similarity measure (\Cref{fig:cross-encoder-setup}).
Finally, we evaluate the model on $\mathcal{K}^\theta_\mathrm{test}$ by measuring its accuracy in retrieving the relevant document $d^+$ among $\{d^+, d^-_1,\dots,d^-_m\}$ for a given query $q$.

The task design ensures that the association between knowledge query $q_i$ and gold fact document $d^+_i$ relies solely on the parametric knowledge represented by $x_i \in \mathcal{D}^\theta$. This is because other variables, like text overlap, are randomly sampled from the same distribution for both query and documents.

Thus, the model can only solve the task by utilizing its internal knowledge.
Finetuning on $\mathcal{K}^\theta_\mathrm{train}$ should only trigger the utilization of the parametric knowledge.

\begin{figure*}[t]
    \begin{subfigure}[t]{0.495\textwidth}
        \centering
        \includegraphics[width=\linewidth]{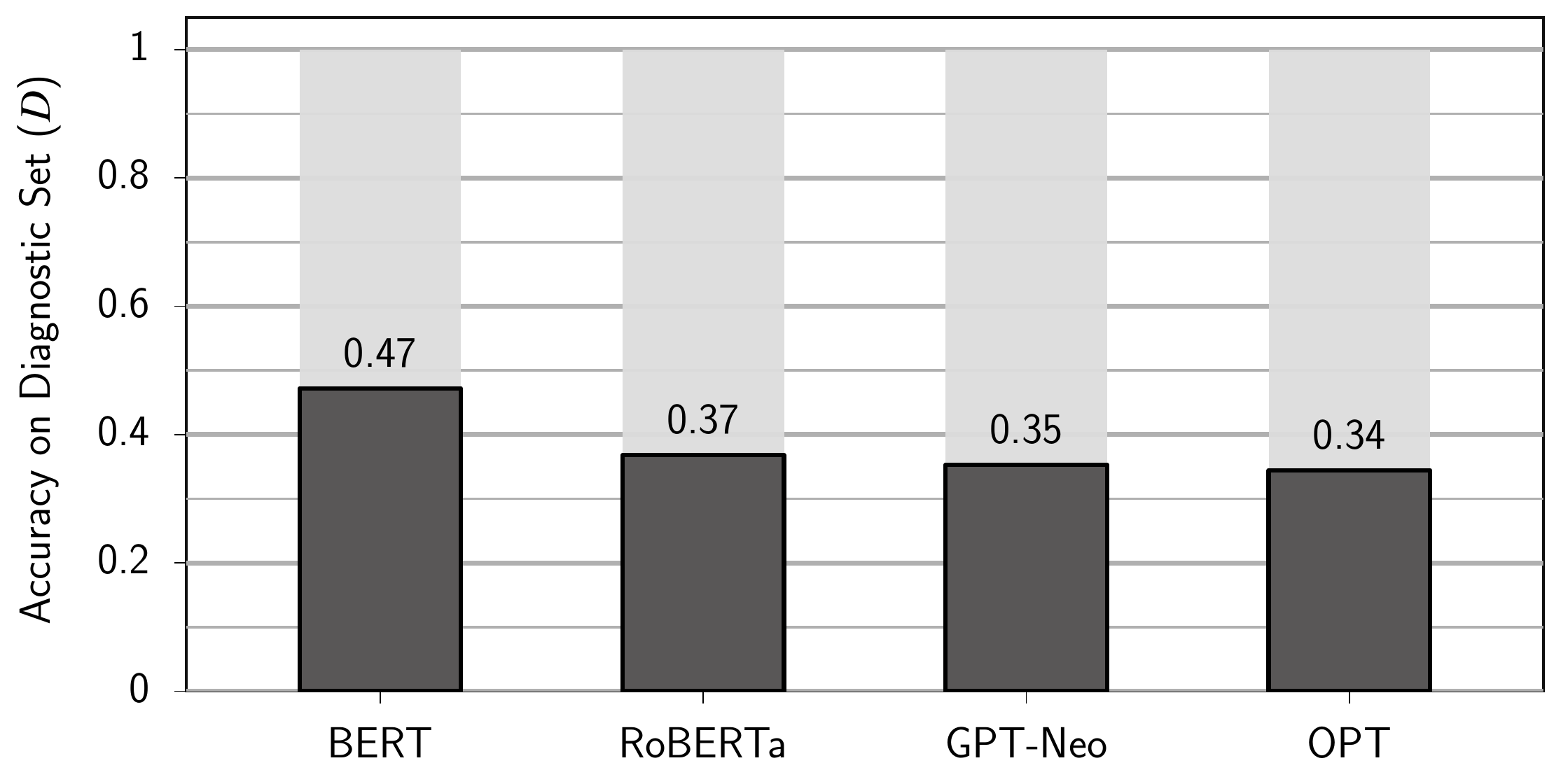}
        \caption{Encyclopedic Knowledge (Zero-shot)}
        \label{fig:param_know_vs_downstream:stage_1}
    \end{subfigure}
    \begin{subfigure}[t]{0.5\textwidth}
        \centering
        \includegraphics[width=\linewidth]{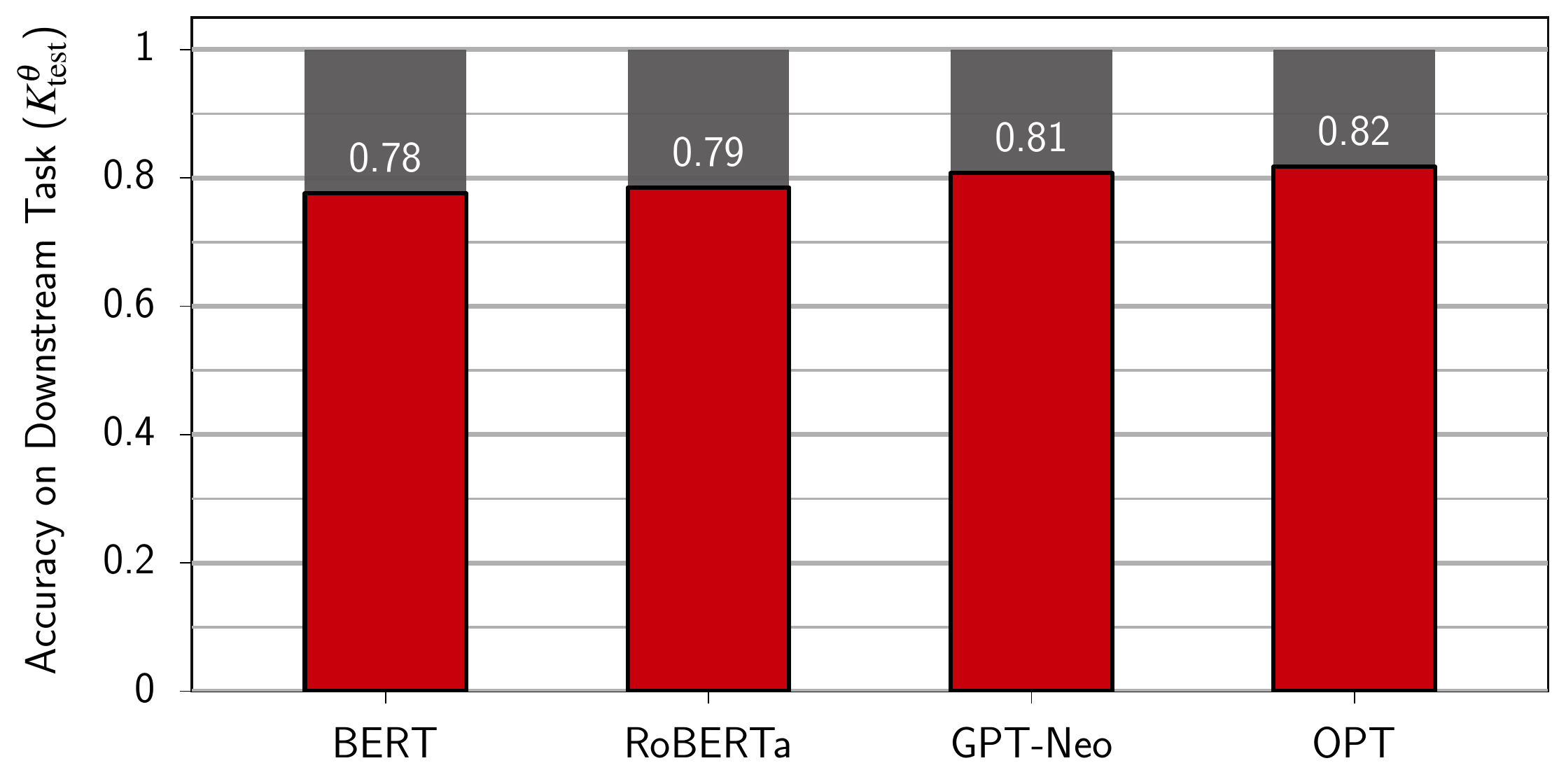}
        \caption{Knowledge Utilization in Downstream (Finetuned)}
        \label{fig:param_know_vs_downstream:stage_2}
    \end{subfigure}
    \caption{
        \textbf{(a)} The fraction of encyclopedic facts the pretrained LM can predict correctly without any training. Reported over three seeds (standard deviation $\sigma \le 0.004$ for all models).
        \textbf{(b)} The model performance in downstream task (created based on correctly predicted facts) measured as top-$1$ retrieval accuracy. Averaged over $27$ runs ($\sigma \le 0.011$ for all models).
    }
    \label{fig:param_know_vs_downstream}
\end{figure*}

\paragraph{Training}
The document generator $\mathrm{P}(d\mid \cdot)$ can generate various types of documents for each fact $x \in \mathcal{D}^\theta$. Please refer to \Cref{tab:document-types} for a list of all the types.
For training, we use three types for \emph{negative} documents $d^-$'s with uniform weights:
$(\headgold{h}, \relgold{r}, \cdot )$,
$(\cdot, \relgold{r}, \tailgold{t})$,
and $(\headgold{h}, \cdot, \tailgold{t})$ as they are the hardest ones since they only differ in one entity from the query.
To keep the GPU memory usage under control, we sample four documents per each type (Refer to \Cref{sec:effect_of_initial_knowledge} for the effect of the number of negatives on the results), which results in total of $12$ negatives.
But, we resample the documents on each epoch to avoid overfitting and use a validation set to choose the best checkpoint.
Also, we keep the learning rate low and use no weight decay to prevent any forgetting.
We use three seeds for the extraction phase, three seeds for splitting $\mathcal{D}^\theta$ into train and test, and three seeds for finetuning on the downstream task,
which results in $27$ different runs per each model.
Refer to \Cref{sec:training-details} for the full list of all hyperparameters)

\paragraph{Inference}
During inference, the model has to recover the gold document $d^+$ from \emph{distractor} documents $d^-$'s.
We use all non-gold document types produced in \Cref{tab:document-types} as distractors(with uniform weights)
and we sample $50$ documents per each,
except for $(\headgold{h}, \relgold{r}, \cdot)$ for which instead of sampling, we enumerate all such documents
to make sure the model actually knows the correct answer.
Furthermore, we include same number of $(\cdot, \cdot, \cdot)$ that are factually correct but not related to the given query. We sample these documents from the test set.

\noindent We evaluate pre-trained models across many families:
OPT \citep{Zhang2022:OPT}, GPT-Neo \citep{Black2021:GPTNeo}, RoBERTa \citep{Liu2019:Roberta}, and BERT \citep{Devlin2019:BERT}. Unless otherwise stated, we use the \texttt{base} size (125M) of these models. We investigate the scaling behavior of model in \Cref{sec:scaling}.
We initialize the diagnostic dataset $\mathcal{D}$ from LAMA \citep{Petroni2019:LAMA}, which has $34K$ facts over $40$ relations. In total, we perform $1134$ finetuning runs.

\section{Evaluating the Knowledge Utilization}
\label{sec:main_question}

We report the fraction of correctly predicted facts from the diagnostic dataset $\mathcal{D}$ and the downstream task performance in \Cref{fig:param_know_vs_downstream}. These results reveal several findings:

First, we find that, on par with previous work \citep{Qin-eisner2021:SoftPrompt}, there is a significant gap in the encyclopedic facts the models can correctly predict and the entire facts present in the diagnostic dataset $\mathcal{D}$ (\Cref{fig:param_know_vs_downstream:stage_1}).
Note that one can arbitrarily increase the number of correctly predicted by considering a prediction as correct if the gold entity is among the model's top-$k$ predictions.
However, we only consider $k=1$ to only focus on the facts that the model can confidently predict.
Nonetheless, we find that BERT and RoBERTa extract slightly more encyclopedic facts than GPT-Neo and OPT.

Critically, all models demonstrate a pronounced gap in downstream task performance, or knowledge utilization, (\Cref{fig:param_know_vs_downstream:stage_2}).
This unexpected outcome occurs despite the downstream task being seemingly simple since (1) models are trained and evaluated on examples based on their accurate encyclopedic knowledge predictions, and (2) both $\mathcal{K}^\theta_\mathrm{train}$ and $\mathcal{K}^\theta_\mathrm{test}$ are sampled from the same distributions (I.I.D), so the models only encounter seen entities.
Notably, OPT and GPT-Neo manage to outperform BERT and RoBERTa by a small margin.

This finding suggests that models struggle to utilize their entire parametric knowledge in downstream tasks. In the next sections, we investigate the potential causes of this gap.

\subsection{Role of Downstream Training Data}
\label{sec:effect_of_initial_knowledge}

\begin{figure*}[t]
    \begin{subfigure}[t]{0.5\textwidth}
        \centering
        \caption{The effect of $|\mathcal{D}^\theta|$ on knowledge utilization}
        \includegraphics[width=\linewidth]{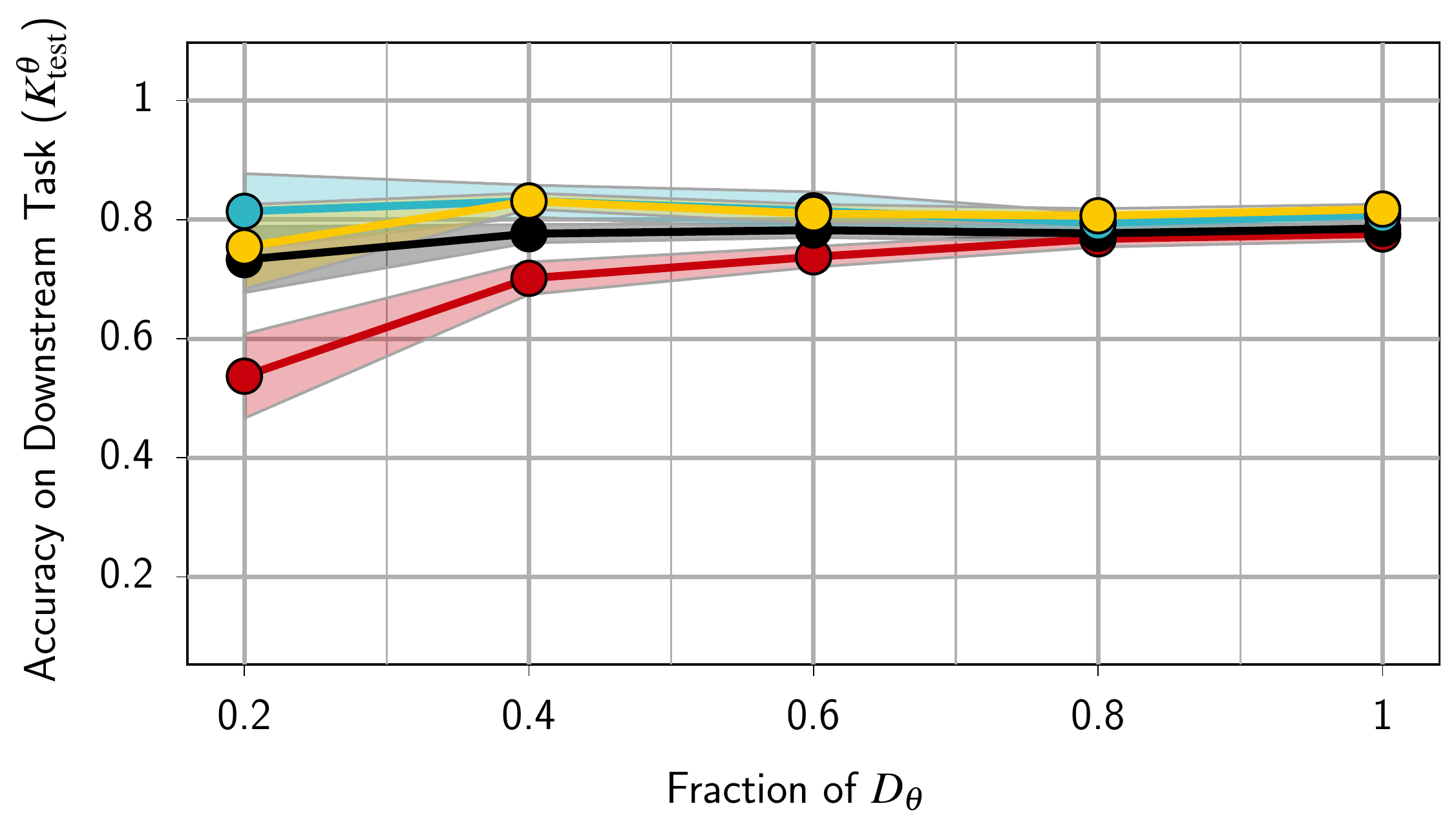}
        \label{fig:data_effect:initial_knowledge}
        \vspace{-10mm}
    \end{subfigure}
    \begin{subfigure}[t]{0.5\textwidth}
        \centering
        \caption{The effect of negative documents on knowledge utilization}
        \includegraphics[width=\linewidth]{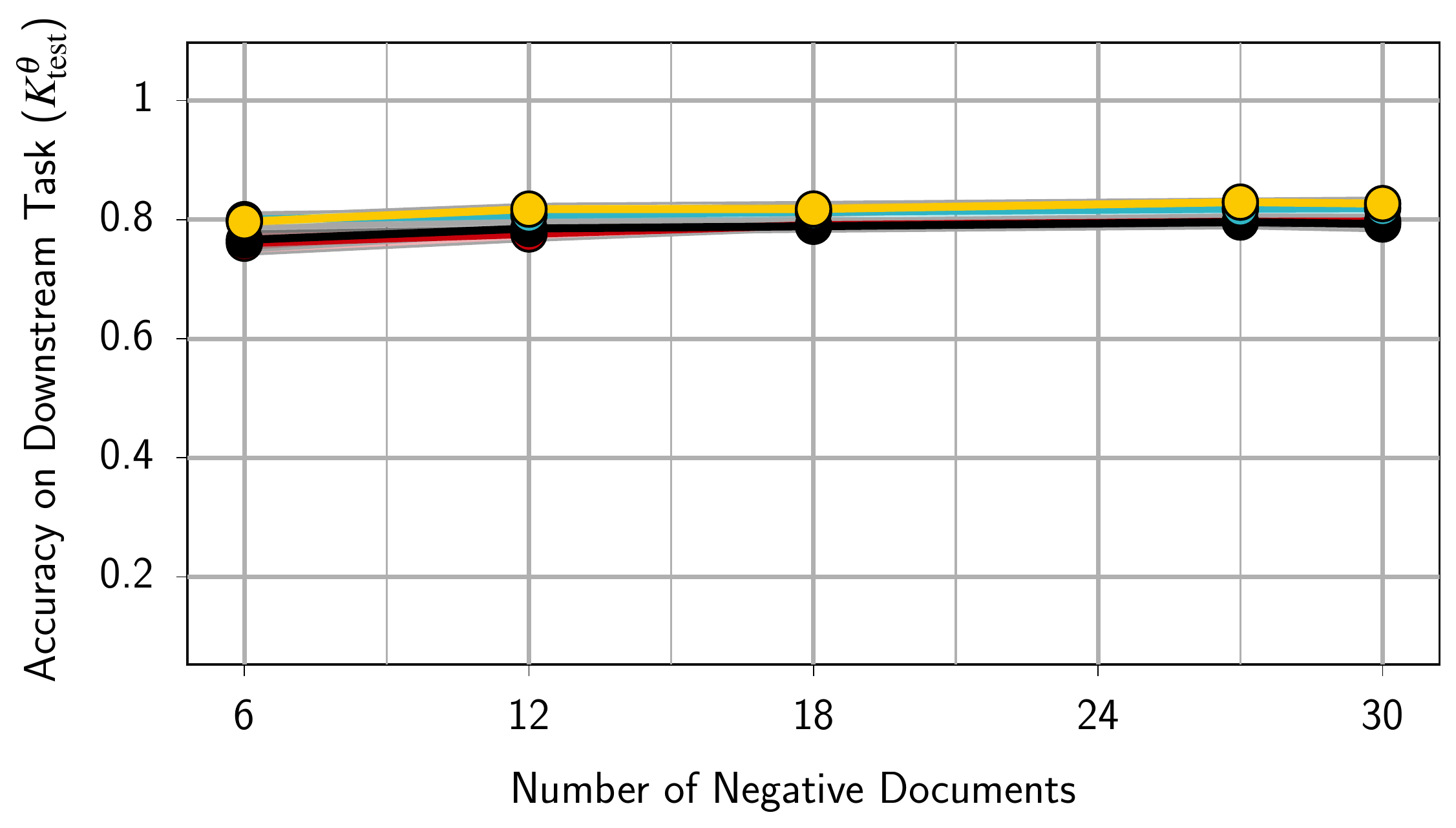}
        \label{fig:data_effect:num_negs}

        \vspace{-10mm}
    \end{subfigure}
    \begin{subfigure}[t]{\textwidth}
        \centering
        \includegraphics[width=\linewidth]{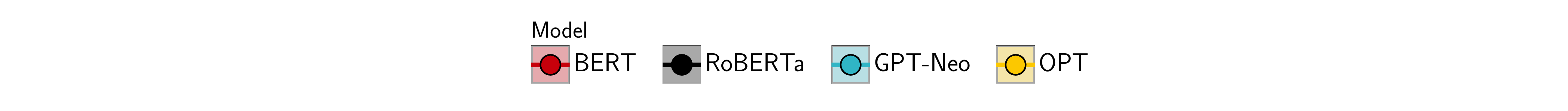}
    \end{subfigure}
    \caption{
        \textbf{(a)} Knowledge utilization when using different fractions of parametric knowledge to create the downstream task.
        \textbf{(b)} The effect of number of negative training documents ($d^-$) used for creating the downstream task.
    }
    \label{fig:data_effect}
\end{figure*}

\paragraph{The effect of initial knowledge $\mathcal{D}^\theta$} As we utilize $\mathcal{D}^\theta$ to create the downstream task, examining the impact of its size ($|\mathcal{D}^\theta|$) on knowledge utilization is crucial. If consistent behavior is observed across different knowledge sizes, it implies that the gap stems from inductive biases (e.g., the model or fine-tuning process), rendering downstream task accuracy a dependable gauge of knowledge utilization.

To measure such effect, for each model, we first compute $\mathcal{D}^\theta$, and then instead of directly using it for $\mathcal{K}^\theta$, we sub-sample smaller sets of it at various fractions and construct the downstream task using each sub-sampled $\mathcal{D}^\theta$.
In \Cref{fig:data_effect:initial_knowledge}, we observe the knowledge utilization is fairly consistent (at least for fractions $> 0.4 $) across different sizes of $\mathcal{D}^\theta$ for all models. Larger fractions seem to have less variance as well. 
This suggests that the utilization performance is intrinsic to the downstream knowledge transfer rather than the initial knowledge residing in the model.

\paragraph{The effect of the number of negatives} The model learns to apply its parametric knowledge by optimizing the retrieval objective. To ensure that the training signal, produced by the contrastive loss on $\mathcal{K}^\theta_\mathrm{train}$, is strong enough, we vary the number of negative documents used for creating $\mathcal{K}^\theta_\mathrm{train}$. If the training signal is weak, we expect the knowledge utilization to improve as we increase the number of negatives.

To answer this question, we simply follow the same setup as described in \Cref{sec:framework} and increase the number of negative documents sampled per each type from $4$ to $10$. We also consider reducing it to two negatives per type to better understand its effectiveness. We keep the initial knowledge $\mathcal{D}^\theta$ fixed. 

\Cref{fig:data_effect:num_negs} summarizes our findings. Knowledge utilization remains the same for all models as we increase the number of negatives. This pattern is observed even when using as few negatives as two per each type. This suggests that the training signal is strong enough across the board and the gap in knowledge utilization is not rooted in the training objective.

\subsection{Gap 1 vs. Gap 2}
\begin{figure}[!h]
    \centering
    \includegraphics[width=\linewidth]{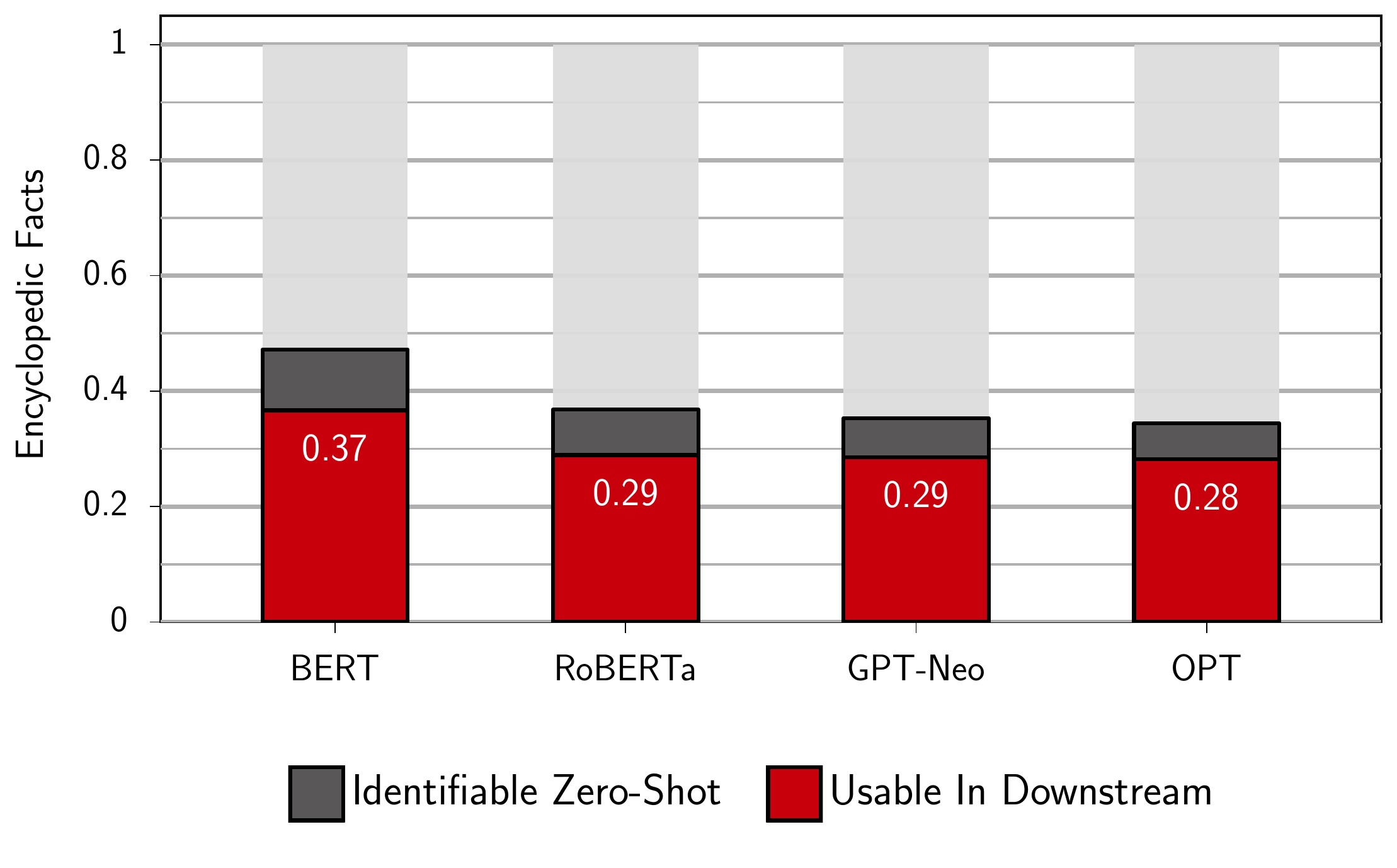}
    \caption{
        \textbf{Gaps in parametric knowledge} \ \ \
        \pill[ark-light-gray]{\phantom{A}} Gap 1 represents the missing facts in parametric
        knowledge $\mathcal{D}^\theta$ (what the model knows).
        \pill[ark-gray]{\phantom{A}} Gap 2 exists in how many of the known facts the model can actually utilize in downstream tasks (the usable knowledge).
    }
    \label{fig:model_knowledge}
\end{figure}

Findings in \Cref{sec:effect_of_initial_knowledge}%
shows that the gap in knowledge utilization (i.e. accuracy on $\mathcal{K}^\theta_\mathrm{test}$) does not depend on the size of $\mathcal{D}^\theta$ and is fairly consistent across different number of negatives.
Moreover, we find that the variation across the random splitting of $\mathcal{D}^\theta$ to create train and test sets of the downstream task is negligible.

The robustness to such design choices allows us to define \emph{Usable Knowledge}, which basically indicates the portion of facts from $\mathcal{D}$ that the model can \emph{actually} utilize in the downstream task. We compute this metric by multiplying the accuracy on $\mathcal{K}^\theta_\mathrm{test}$ by the fraction of correctly predicted facts in $\mathcal{D}$. We report the results in \Cref{fig:model_knowledge}.

These results clearly demonstrate that there exist two gaps in the models' knowledge.
Gap $1$ is in how many facts  the model knows after pre-training.
Gap $2$ is in how many of facts the model knows can be truly utilized in downstream tasks.

Indeed, we see that although RoBERTa manages to extract more facts than GPT-Neo, due to Gap $2$, it performs the same as GPT-Neo in downstream tasks.

\section{Robustness of Knowledge Utilization}
\label{sec:robustness}
We intentionally design the downstream task $\mathcal{K}^\theta$ to be straightforward and free of any distributional shift as we want to measure the \emph{maximum} knowledge utilization of the model. 
However, in real-world applications, it is likely that the model encounter samples that are different from the training distribution.
In this section, we investigate the robustness of knowledge application in the presence of such distributional shifts.

\subsection{Non-I.I.D. $\mathcal{K}^\theta_\mathrm{train}$ and $\mathcal{K}^\theta_\mathrm{test}$}
Recall that we randomly divide $\mathcal{D}^\theta$ into two sets as the data source for the creation of $\mathcal{K}^\theta_\mathrm{train}$ and $\mathcal{K}^\theta_\mathrm{test}$. 
In this experiment, however, we split $\mathcal{D}^\theta$ such that the relation types ($\mathbf{r}$) in $\mathcal{K}^\theta_\mathrm{train}$ and $\mathcal{K}^\theta_\mathrm{test}$ are disjoint. Specifically, we randomly select 60\% of the relations and their corresponding facts for $\mathcal{K}^\theta_\mathrm{train}$ and the rest for $\mathcal{K}^\theta_\mathrm{test}$. We repeat this process over three seeds to create three different splits for $\mathcal{K}^\theta$.
We still follow the same procedure for converting knowledge triples to document retrieval examples as explained in \Cref{sec:framework}.
In this way, we make sure we do not change the nature of the task, i.e. the model still needs to apply its parametric knowledge to solve the task, but the distributional shift between $\mathcal{K}^\theta_\mathrm{train}$ and $\mathcal{K}^\theta_\mathrm{test}$ can potentially represent real-world scenarios. If the model learns to systematically apply its knowledge, we expect its downstream performance to be similar to or close to the I.I.D. setting (\Cref{sec:main_question}).

\begin{figure}[t]
    \centering
    \includegraphics[width=\linewidth]{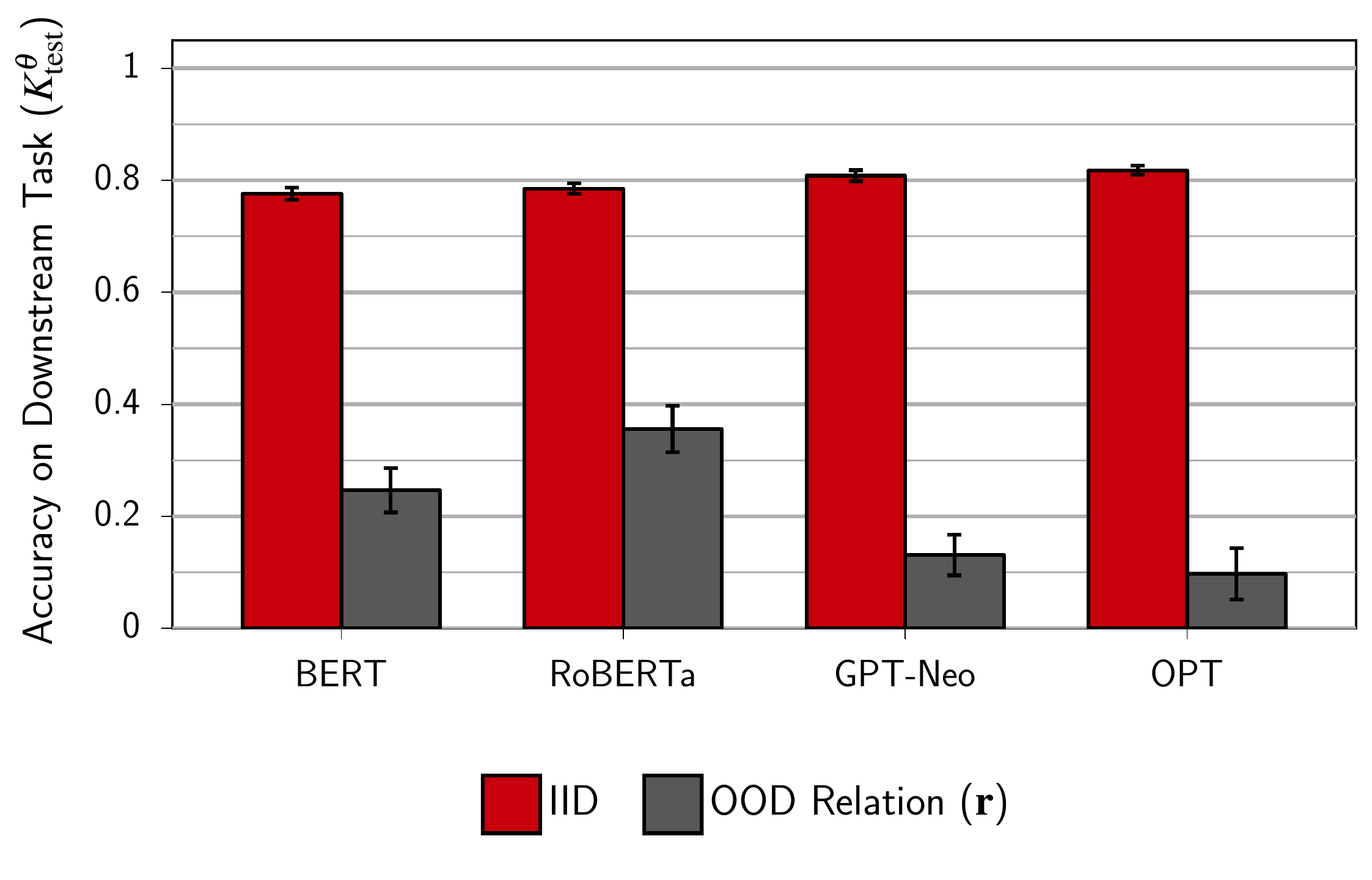}
    \caption{
        \textbf{Robustness to distributional shift} \ \ \
        In the OOD setting, we produce a distributional shift (over the relation types) between the examples in the train and test set of the downstream task $\mathcal{K}^\theta$. All models fail to generalize to unseen relations. The IID setting is the same as the one described in \Cref{sec:framework} and repeated from \Cref{fig:param_know_vs_downstream:stage_2} for comparison.
    }
    \label{fig:ood_rel}
\end{figure}

We observe downstream task performance drops significantly for all models when evaluated OOD (\Cref{fig:ood_rel}). This indicates the models cannot use their knowledge on examples with unseen relation types, though all relations and facts originate in $\mathcal{D}^\theta$. Thus, knowledge usage in downstream tasks is sensitive to distribution shifts, suggesting failure to apply pretraining knowledge may be more severe in real-world applications.

\section{Effect of Scaling law On The Gaps}
\label{sec:scaling}

Recent NLP success has come from scaling up pretraining model parameters \citep{Brown2020:GPT3}. With larger models and increased compute, capabilities such as in-context learning and chain-of-thought reasoning emerge \citep{Wei2022:LLMEmergentAbilites}. 
The expanded capacity allows these models to absorb more knowledge from pretraining data, improving their usefulness as knowledge sources. However, it remains uncertain if scaling boosts the proportion of pretraining knowledge applicable to downstream tasks. Ideally, we like to see a narrowing gap in pretraining knowledge alongside superior knowledge utilization.

\begin{figure}[!t]
    \centering
    \includegraphics[width=\linewidth]{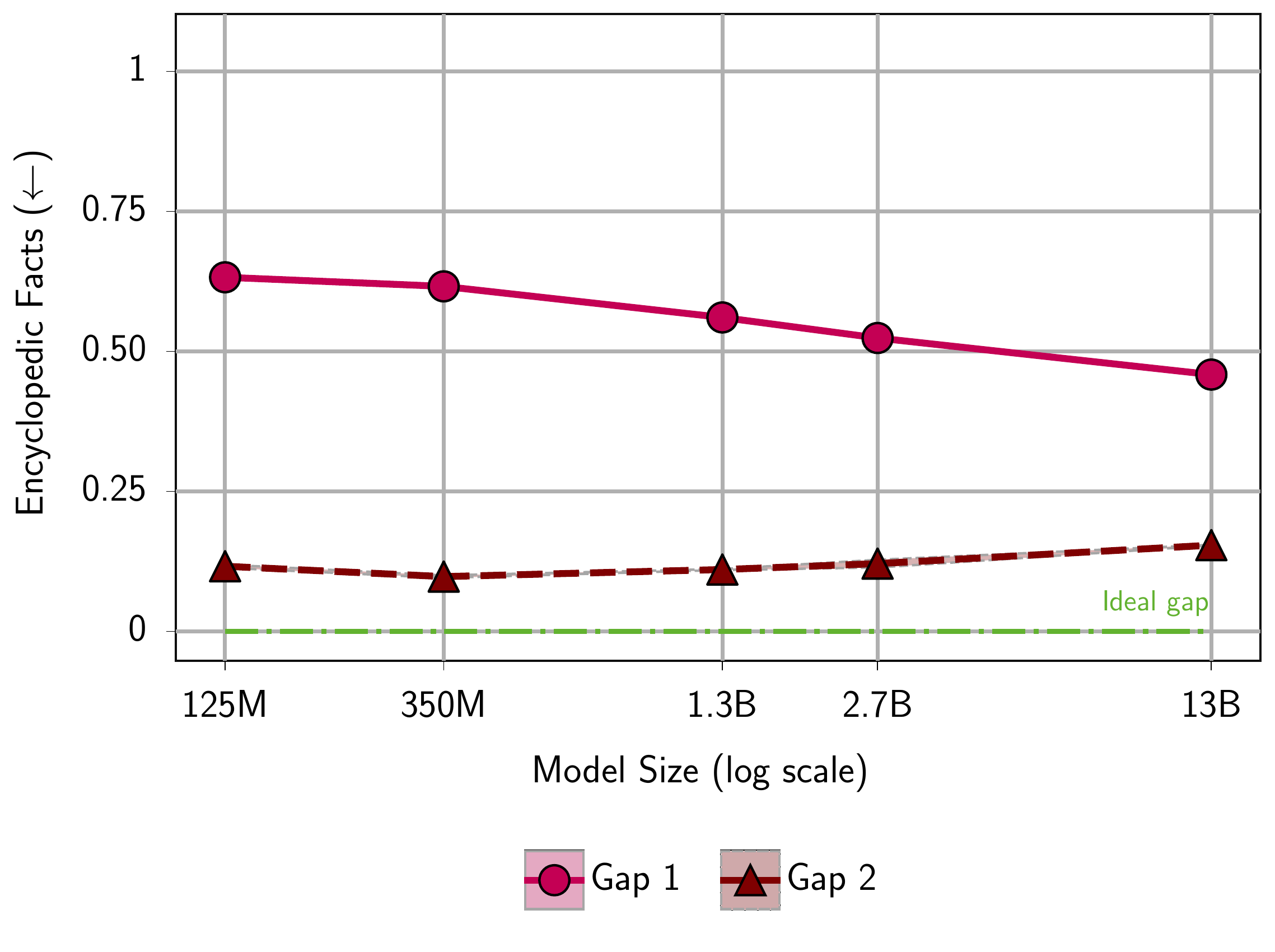}
    \caption{
        \textbf{Gaps in parametric knowledge} \ \ \
        Knowledge gaps directly compute across different model sizes. Specifically, we use $1 - (\text{Accuracy on } \mathcal{D}^\theta)$ for Gap 1 and $(\text{Accuracy on } \mathcal{D}^\theta) \times (1 - (\text{downstream accuracy})$) for Gap 2.  
    }
    \label{fig:gaps_in_scale}
\end{figure}

To investigate this, we evaluate \methodabbrev on increasing sizes of OPT \citep{Zhang2022:OPT}. Specifically, at each  scale, we first extract the model's parametric knowledge and then create the downstream task based on it using the same procedure as described in \Cref{sec:framework}. \Cref{fig:teaser_scaling_acc} reports the results of this experiment. We observe the following trends:

First, we confirm that a greater fraction of knowledge triples in $\mathcal{D}$ can be identified in larger models, suggesting they acquire more knowledge from pretraining data. Secondly, we find that the gap between identifiable and usable knowledge persists in larger models, and their ability to apply knowledge in downstream tasks does not improve with scaling. \Cref{fig:gaps_in_scale} illustrates these gaps directly, demonstrating that while Gap $1$ decreases in larger models, Gap $2$ remains relatively unchanged.

The results suggest that while PLMs, even at small scales, pose considerable knowledge, extracting an equivalent amount of usable knowledge necessitates much larger models. For instance, OPT-125M accurately predicts $34$\% of encyclopedic facts, but only OPT-13B (approximately $100\times$ larger) can reliably apply the same volume in downstream tasks. An ideal model should address both issues and from \Cref{fig:gaps_in_scale} it is justifiable %
that %
having a higher amount of parametric knowledge does not guarantee improved downstream performance.

\section{Discussion}
Lately, pretrained language models with chatbot interfaces have increasingly served as knowledge bases \citep{Ouyang2022:InstructGPT}. These chatbots typically employ the model's parametric knowledge to respond to queries and offer information. Our study examines the dependability of this knowledge and its impact on downstream task performance. We discover that, regardless of inductive biases, PLMs face difficulty utilizing their full knowledge in downstream tasks (\Cref{sec:main_question}). This unreliability of parametric knowledge could constrain the concept of ``PLMs as differentiable knowledge bases.''

Additionally, our findings show that the utilization gap persists even with scaling (\Cref{sec:scaling}). Notably, while models at each scale capture more knowledge from pretraining data, obtaining the same amount of usable knowledge requires significantly larger models. This exposes a potential constraint in the recent trend of adopting mid-sized PLMs \citep{Li2023:Starcoder}.

Lastly, we discover that knowledge utilization depends on the peculiarities of fine-tuning data for downstream tasks.
Specifically, as seen in \Cref{sec:robustness}, PLMs struggle to apply their knowledge to relation types not encountered during fine-tuning, even if they accurately predicted such facts in step $1$. This generalization gap could highlight challenges within the recent SFT-RLHF paradigm \citep{Ouyang2022:InstructGPT}. For instance, the model may only adhere to instructions and excel at tasks resembling the fine-tuning data. Consequently, it might be necessary to meticulously craft fine-tuning data to activate and utilize all aspects of parametric knowledge in downstream tasks.
However, it requires elaborate studies to establish the systematic issues in knowledge application beyond encyclopedic knowledge like procedural and task knowledge.

\section{Related Work}
\paragraph{Parametric Knowledge}
\citet{Petroni2019:LAMA} constructed a probing dataset to measure the factual knowledge of present in PLMs.
They showed that many encyclopedic facts can be extracted without further training of the model and proposed PLMs as a new type of knowledge base, which can be trained on the unstructured text and queried using natural language.
Follow-up work improves the methods for probing and extracting world knowledge from PLMs \citep{Jiang2020:LPAQA,Shin2020:AutoPrompt,Qin-eisner2021:SoftPrompt,Newman2022:padapters}.
Apart from encyclopedic facts, studies have explored PLMs' parametric knowledge in other areas, such as linguistic structures \citep{Tenney2018:LearnContext,Blevins2022:PromptingLing}, and commonsense \citep{Zhou2020:CommonSense, Li2022:ComSenseQAExtractedKnow}. Recently, the emergent abilities of LLMs have shown that they acquire skills like coding \citep{Chen2021:Codex}, reasoning \citep{Chowdhery2022:PaLM}, and in-context learning \citep{Brown2020:GPT3}, in addition to the previously mentioned knowledge.

\paragraph{Using the Parametric Knowledge}
\citet{Roberts2020:T5ClosedQA} finetune a pretrained T5 model for question answering in a closed-book setting and showed that it can perform on par or better than models that use explicit knowledge bases. \citet{Wang2021:BartClosedQA} made a similar observation for the BART model.
More recently, PLMs are being used to generate facts and documents for knowledge-intensive tasks \citep{Li2022:DialogKnowledgeExtraction,Liu2022:MultiStagePrompt,Yu2023:GenRead}. In this paradigm, in order to answer factual questions, instead of retrieving relevant documents, 
the model has to first generate the facts and then answer the question with those facts as context.
This paradigm shows that the models may not be able to use their parametric knowledge on their own and need explicit grounding to be able to use it.
Furthermore, there is a plethora of work that investigates whether the model employs its linguistic knowledge when solving downstream language understanding tasks.
\citet{Mccoy2019:RightForWrong} shows that RoBERTa does not use its linguistic knowledge for solving NLI. Instead, it relies on shallow heuristics. \citet{Lovering2021:PredictingInductiveBias}'s observation aligns with this finding and shows the training data used for the downstream task needs to have enough evidence to trigger the model's linguistic knowledge.
In our work, we use a more general notation of parametric knowledge and investigate utilization in cases where sufficient evidence is present in the finetuning data.

\section{Conclusion}

In this study, we presented \methodname (\methodabbrev), a framework designed to assess the parametric knowledge of pretrained language models. Employing \methodabbrev, we identified a previously unnoticed gap in what models know and how much of it they can actually use. Our findings reveal that this gap exists not only in smaller models but also persists in larger ones. Additionally, we demonstrate that a distributional shift in fine-tuning data can result in even larger gaps between the model's knowledge and its practical application in downstream tasks.

\bibliography{anthology,custom}
\bibliographystyle{acl_natbib}

\appendix

\renewcommand\thefigure{\thesection.\arabic{figure}}
\setcounter{figure}{0}

\section{Training Details}
\label{sec:training-details}

\subsection{Knowledge Extraction}
We follow the same procedure as \citet{Qin-eisner2021:SoftPrompt} to extract knowledge facts from a frozen PLM.
Specifically, we use soft-prompt instead of discrete prompts. We append three soft-prompts before and after the head entity.
Assign different soft-prompts per each relation type. Finally, we train them using the train-set provided by \citet{Petroni2019:LAMA} and use a validation set to select the best checkpoint. \Cref{tab:hyperparam:knowledge_extraction} summarizes the hyperparameters used in this stage, which we borrow from \citet{Qin-eisner2021:SoftPrompt}.

\subsection{Finetuning}
We follow a straightforward procedure for finetuning the models. \Cref{tab:hyperparam:finetuning} lists the hyperparameters we used for finetuning. In the initial experiments, we tried $\mathrm{lr} \in \{\num{1e-5}, \num{3e-5}, \num{5e-5}\}$ however, we did not find any significant difference between them for all models. Therefore, we decided to use the same learning rate for all models.

\begin{table}[h]
    \centering
    \vspace{3mm}
    \footnotesize
    \begin{tabular}{
            l@{\hskip 0.2in}
            r
        }
        \toprule
        Parameter               & Value                 \\
        \midrule
        Optimizer               & AdamW                 \\
        Learning rate           & \num{1e-4}                \\
        Weight Decay            & 0                     \\
        Batch size              & 64                    \\
        Learning Rate Scheduler & Polynomial            \\
        Warm Up                 & 6\% of training steps \\
        \# Train Epochs         & 20                    \\
        \bottomrule
    \end{tabular}
    \caption{Summary of hyperparameters used in knowledge extraction stage (stage 1).}
    \label{tab:hyperparam:knowledge_extraction}
\end{table}{}

\begin{table}[h]
    \centering
    \vspace{3mm}
    \footnotesize
    \begin{tabular}{
            l@{\hskip 0.2in}
            r
        }
        \toprule
        Parameter               & Value                 \\
        \midrule
        Optimizer               & AdamW                 \\
        Learning rate           & \num{1e-5}               \\
        Weight Decay            & 0                     \\
        Batch size              & 32                    \\
        Learning Rate Scheduler & Polynomial            \\
        Warm Up                 & 6\% of training steps \\
        \# Train Epochs         & 20                    \\
        \bottomrule
    \end{tabular}
    \caption{Summary of hyperparameters used in finetuning on downstream task (stage 2).}
    \label{tab:hyperparam:finetuning}
\end{table}{}

\end{document}